\pdfoutput=1

\documentclass[11pt]{article}

\usepackage{ACL2023}

\usepackage{times}
\usepackage{latexsym}

\usepackage[T1]{fontenc}

\usepackage[utf8]{inputenc}

\usepackage{microtype}

\usepackage{amsfonts}
\usepackage{amsmath}
\usepackage{booktabs}
\usepackage{graphicx}
\usepackage{comment}
\usepackage{todonotes}
\usepackage{multirow}
\usepackage{xspace}

\newcommand{\ourmodelfull}{\texttt{Chain-of-Skills (COS)}\xspace}
\newcommand{\ourmodellong}{\texttt{Chain-of-Skills}\xspace}
\newcommand{\ourmodelshort}{\texttt{COS}\xspace}
\input{Definitions.tex}

\newcommand{\clstoken}{\texttt{[CLS]}\xspace}
\newcommand{\septoken}{\texttt{[SEP]}\xspace}

\newcommand\blfootnote[1]{%
  \begingroup
  \renewcommand\thefootnote{}\footnote{#1}%
  \addtocounter{footnote}{-1}%
  \endgroup
}
\title{Chain-of-Skills: A Configurable Model for Open-Domain \\Question Answering}

\author{
Kaixin Ma\textsuperscript{$\clubsuit$}$\dagger^*$,
Hao Cheng\textsuperscript{$\spadesuit$}$^*$,
Yu Zhang\textsuperscript{$\heartsuit$}$\dagger$,
Xiaodong Liu\textsuperscript{$\spadesuit$},
Eric Nyberg\textsuperscript{$\clubsuit$},
Jianfeng Gao\textsuperscript{$\spadesuit$}
 \\ 
  \textsuperscript{$\clubsuit$} Carnegie Mellon University
  \textsuperscript{$\spadesuit$} Microsoft Research \\
  \textsuperscript{$\heartsuit$} University of Illinois at Urbana-Champaign
 \\
  {\small \tt \{kaixinm,ehn\}@cs.cmu.edu} \ 
  {\small \tt \{chehao,xiaodl,jfgao\}@microsoft.com} \ 
  {\small \tt yuz9@illinois.edu} 
}

\hypersetup{draft}
\begin{document}
\maketitle
\begin{abstract}
The retrieval model is an indispensable component for real-world knowledge-intensive tasks, \eg open-domain question answering (ODQA).
As separate retrieval skills are annotated for different datasets, recent work focuses on customized methods, limiting the model transferability and scalability.
In this work, we propose a modular retriever where individual modules correspond to key skills that can be reused across datasets.
Our approach supports flexible skill configurations based on the target domain to boost performance.
To mitigate task interference, we design a novel modularization parameterization inspired by sparse Transformer.
We demonstrate that our model can benefit from self-supervised pretraining on Wikipedia and fine-tuning using multiple ODQA datasets, both in a multi-task fashion.
Our approach outperforms recent self-supervised retrievers in zero-shot evaluations and achieves state-of-the-art fine-tuned retrieval performance on NQ, HotpotQA and OTT-QA.
\blfootnote{$\dagger$ Work done during an internship at Microsoft Research} \blfootnote{$*$ Equal contribution}
\end{abstract}

\begin{figure}[!ht]
    \centering
    \includegraphics[scale=0.38]{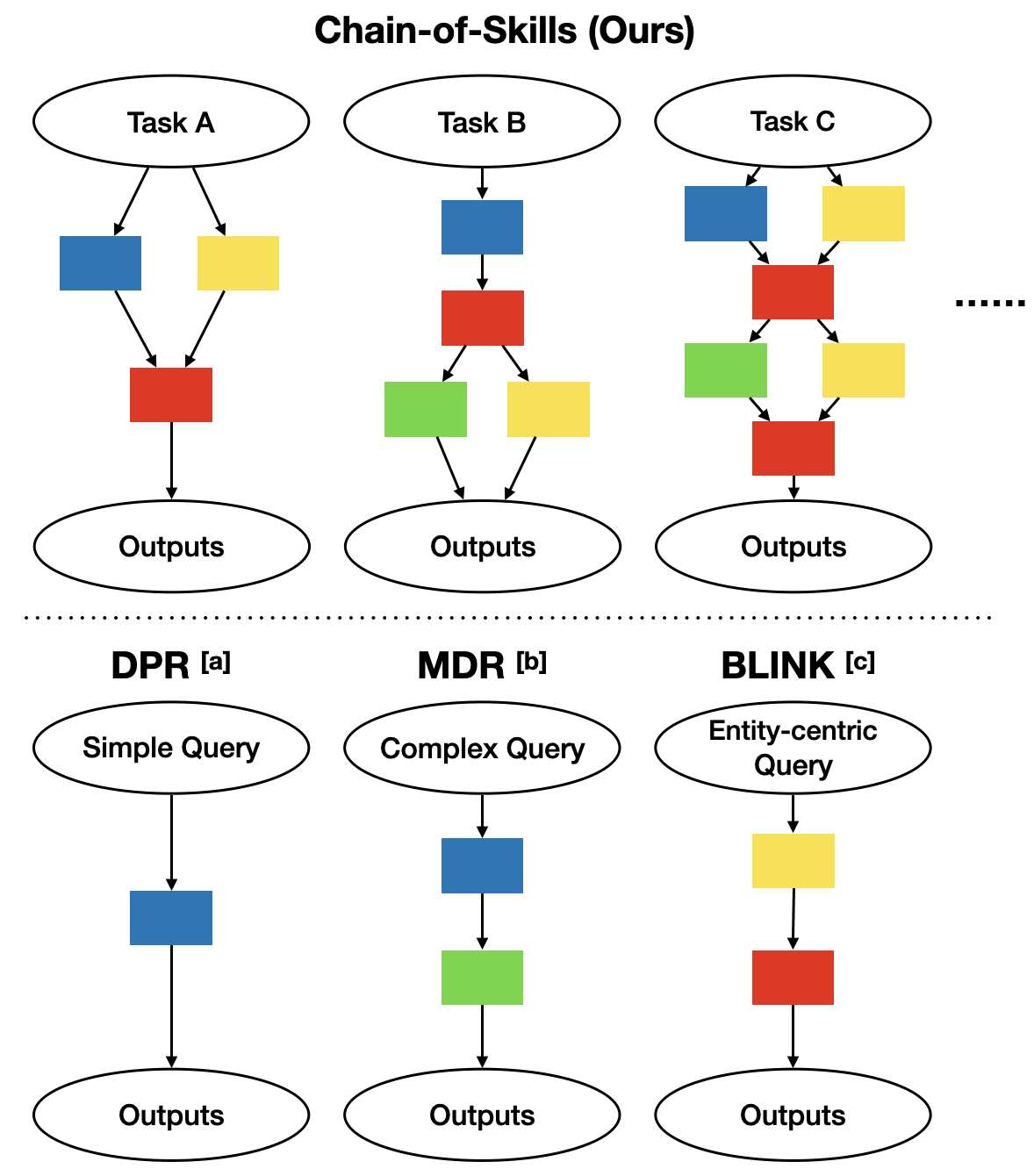}
    \caption{Comparison of dense retrievers in terms of considered query type and supported skill configuration
 $^{[a]}$\cite{karpukhin-etal-2020-dense}
 $^{[b]}$\cite{xiong2020answering} 
 $^{[c]}$\cite{wu-etal-2020-scalable}. Each box represents a skill (
\colorbox{blue}{\color{blue}a}$=$\textit{single retrieval},
\colorbox{green}{\color{green}a}$=$\textit{expanded retrieval},
\colorbox{yellow}{\color{yellow}a}$=$\textit{linking},
\colorbox{red}{\color{red}a}$=$\textit{reranking},
) and the arrows represent the order of execution. In our case, we can flexibly combine and chain the skills at inference time for different tasks to achieve optimal performance.}
    \label{fig:intro}
\end{figure}

\section{Introduction}

Gathering supportive evidence from external knowledge sources is critical for knowledge-intensive tasks, 
such as open-domain question answering \citep[ODQA;][]{lee-etal-2019-latent} and fact verification \cite{thorne-etal-2018-fever}.
Since different ODQA datasets focus on different information-seeking goals, this task typically is handled by customized retrieval models \cite{karpukhin-etal-2020-dense,yang-etal-2018-hotpotqa,wu-etal-2020-scalable,ma2022opendomain}.
However, this dataset-specific paradigm has limited model scalability and transferability.
For example, augmented training with single-hop data hurts multi-hop retrieval \cite{xiong2020answering}.
Further, as new information needs constantly emerge, dataset-specific models are hard to reuse.

In this work, we propose \ourmodellong (\ourmodelshort), a modular retriever based on Transformer \cite{vaswani2017attention},
where each module implements a \textit{reusable} skill 
that can be used for different ODQA datasets.
Here, we identify a set of such retrieval reasoning skills:
\textit{single retrieval, expanded query retrieval, entity span proposal, entity linking} and \textit{reranking} (\S\ref{sec:background}).
As shown in \autoref{fig:intro}, recent work has only explored certain skill configurations.
We instead consider jointly learning all skills in a multi-task contrastive learning fashion.
Besides the benefit of solving multiple ODQA datasets, our multi-skill formulation provides unexplored ways to chain skills for individual use cases.
In other words, it allows flexible configuration search according to the target domain, which can potentially lead to better retrieval performance (\S\ref{sec:exp}).

For multi-task learning, one popular approach is to use a shared text encoder \cite{liu-etal-2019-mtdnn}, \ie sharing representations from Transformer and only learning extra task-specific headers atop.
However, this method suffers from undesirable task interference, \ie negative transfer among retrieval skills.
To address this, we propose a new modularization parameterization inspired by the recent mixture-of-expert in sparse Transformer \cite{fedus2021switch},
\ie mixing specialized and shared representations.
Based on recent analyses on Transformer \cite{meng2022locating}, we design an attention-based alternative that is more effective in mitigating task interference (\S\ref{sec:analysis}).
Further, we develop a multi-task pretraining using \textit{self-supervision} on Wikipedia so that the pretrained \ourmodelshort can be directly used for retrieval without dataset-specific supervision.

To validate the effectiveness of \ourmodelshort, we consider zero-shot and fine-tuning evaluations with regard to the model in-domain and cross-dataset generalization. 
Six representative ODQA datasets are used: Natural Questions \citep[NQ;][]{kwiatkowski-etal-2019-natural}, WebQuestions \citep[WebQ;][]{berant-etal-2013-semantic}, SQuAD \cite{rajpurkar-etal-2016-squad}, EntityQuestions \cite{Sciavolino-etal-2021-simple}, HotpotQA \cite{yang-etal-2018-hotpotqa} and OTT-QA \cite{chen2021ottqa}, where the last two are multi-hop datasets.
Experiments show that our multi-task pretrained retriever achieves superior \textit{zero-shot} performance compared to recent state-of-the-art (SOTA) \textit{self-supervised} dense retrievers and BM25 \cite{10.1561/1500000019}.
When fine-tuned using multiple datasets jointly, \ourmodelshort can further benefit from high-quality supervision effectively, leading to new SOTA retrieval results across the board. 
Further analyses show the benefits of our modularization parameterization for multi-task pretraining and fine-tuning, as well as flexible skill configuration via \ourmodellong inference.\footnote{Data and code available at \url{https://github.com/Mayer123/UDT-QA}}

\begin{figure*}[th!]
    \centering
    \includegraphics[width=0.99\textwidth]{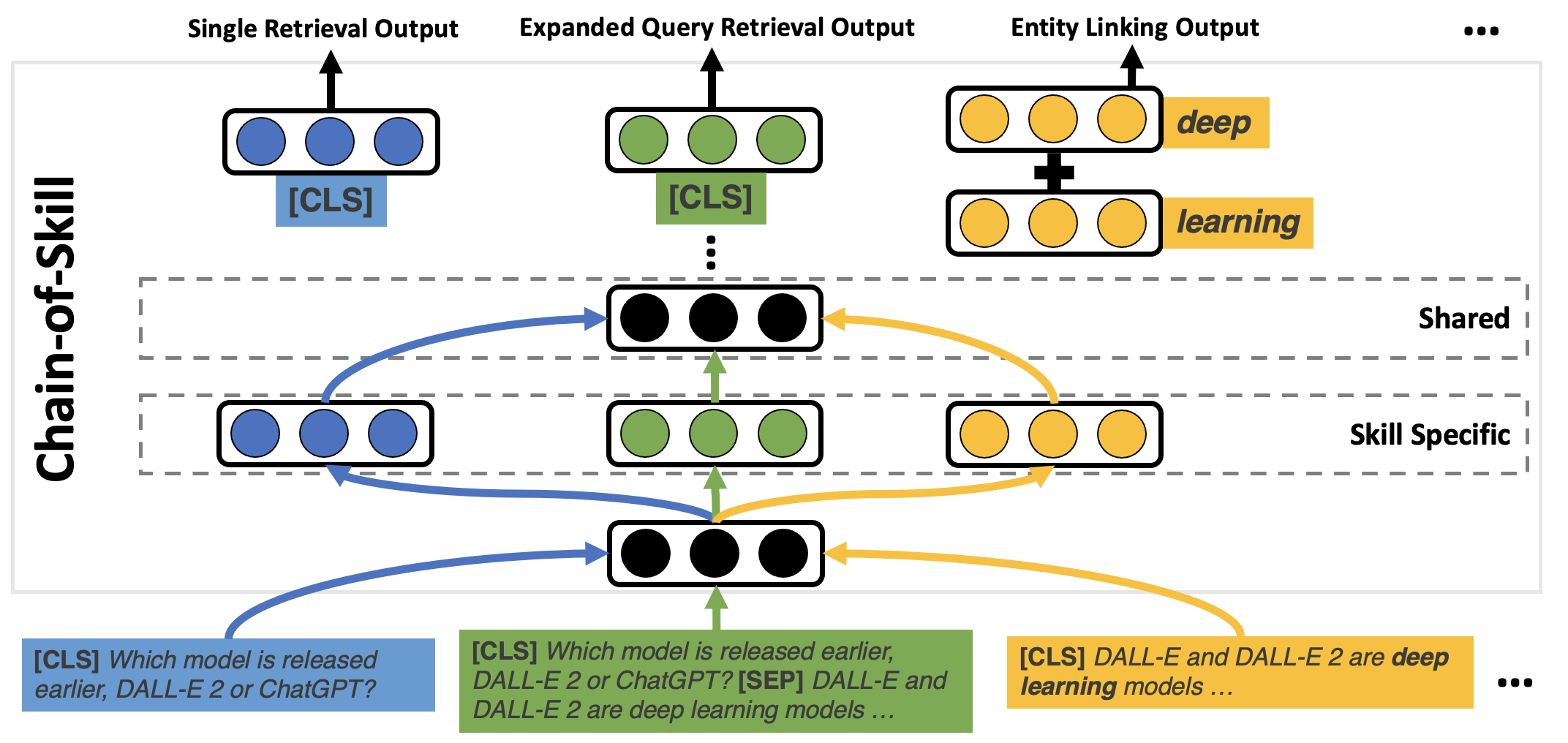}
    \caption{\ourmodelfull model architecture with three different query types. The left blue box indicates the single retrieval query input. The middle green box is the expanded query retrieval input based on the single retrieval results. The right orange case is the entity-centric query with ``deep learning'' as the targeted entity.}
    \label{fig:model_figure}
\end{figure*}

\section{Background}
\label{sec:background}

We consider five retrieval reasoning skills:
\textit{single retrieval}, \textit{expanded query retrieval}, 
\textit{entity linking}, 
\textit{entity span proposal} and \textit{reranking}.
Conventionally, each dataset provides annotations on a different combination of skills (see \autoref{tab:stats}).
Hence, we can potentially obtain training signals for individual skills from multiple datasets.
Below we provide some background for these skills.

\noindent\textbf{Single Retrieval}
Many ODQA datasets \citep[\eg NQ;][]{kwiatkowski-etal-2019-natural} concern simple/single-hop queries.
Using the original question as input (\autoref{fig:model_figure} bottom-left),
single-retrieval gathers isolated supportive passages/tables from target sources in one shot \cite{karpukhin-etal-2020-dense}.

\noindent\textbf{Expanded Query Retrieval}
To answer complex multi-hop questions ,
it typically requires evidence chains of two or more separate passages \citep[\eg HotpotQA;][]{yang-etal-2018-hotpotqa} or tables \citep[\eg OTT-QA;][]{chen2021ottqa}.
Thus, follow-up rounds of retrieval are necessary after the initial single retrieval.
The expanded query retrieval \cite{xiong2020answering} takes an expanded query as input, where the question is expanded with the previous-hop evidence (\autoref{fig:model_figure} bottom-center).
The iterative retrieval process generally shares the same target source.

\noindent\textbf{Entity Span Proposal}
Since many questions concern entities, 
detecting those salient spans in the question or retrieved evidence is useful.
The task is related to named entity recognition (NER), except requiring only binary predictions, \ie whether a span corresponds to an entity.
It is a prerequisite for generating entity-centric queries (context with target entities highlighted; \autoref{fig:model_figure} bottom-right) where targeted entity information can be gathered via downstream entity linking.

\noindent\textbf{Entity Linking}
Mapping detected entities to the correct entries in
a database is crucial for analyzing factoid questions.
Following \citet{wu-etal-2020-scalable}, we consider an entity-retrieval approach, \ie using the entity-centric query for retrieving its corresponding Wikipedia entity description.

\noindent\textbf{Rereanking}
Previous work often uses a reranker to improve the evidence recall in the top-ranked candidates.
Typically, the question with a complete evidence chain is used together for reranking.

\section{Approach}
In this work, we consider a holistic approach to gathering supportive evidence for ODQA, \ie the evidence set contains both singular tables/passages (from single retrieval) and connected evidence chains (via expanded query retrieval/entity linking).
As shown in \autoref{fig:model_figure}, \ourmodelshort supports flexible skill configurations, \eg expanded query retriever and the entity linker can build upon the single-retrieval results.
As all retrieval skill tasks are based on contrastive learning, we start with the basics for our multi-task formulation.
We then introduce our modularization parameterization for reducing task interference.
Lastly, we discuss ways to use self-supervision for pretraining and inference strategies.

\subsection{Reasoning Skill Modules}
\label{ssec:reasoning_skill_modules}

All reasoning skills use text encoders based on Transformer \cite{vaswani2017attention}.
Particularly, only BERT-base \cite{devlin-etal-2019-bert} is considered without further specification. 
Text inputs are prepended with a special token \clstoken~and different segments are separated by the special token \septoken.
The bi-encoder architecture \cite{karpukhin-etal-2020-dense} is used for single retrieval, expanded query retrieval, and entity linking. We use dot product for sim$(\cdot, \cdot)$.

\noindent \textbf{Retrieval} 
As single retrieval and expanded query retrieval only differ in their query inputs, these two skills are discussed together here.
Specifically, both skills involve examples of a question $Q$, a positive document $P^+$.
Two text encoders are used, \ie a query encoder for questions and a context passage encoder for documents.
For the expanded query case (\autoref{fig:model_figure} bottom-center), 
we concatenate $Q$ with the previous-hop evidence
as done in \citet{xiong2020answering}, \ie 
\clstoken~$Q$ \septoken~$P^+_1$ \septoken.
Following the literature, \clstoken~vectors from both encoders are used to represent the questions and documents respectively.
The training objective is
\begin{equation}
      L_\textsubscript{ret} = -{\exp(\text{sim}(\qvec, \pvec^+)) \over \sum_{\pvec^\prime\in\mathcal{P}\cup\{\pvec^+\}} \exp(\text{sim}(\qvec, \pvec^\prime))},
    \label{eqn:obj_retrieve}  
\end{equation}
where $\qvec, \pvec$ are the query and document vectors respectively and $\mathcal{P}$ is the set of negative documents. 

\noindent \textbf{Entity Span Proposal}
To achieve a multi-task formulation, we model entity span proposal based on recent contrastive NER work \cite{zhang2022optimizing}. 
Specifically, for an input sequence with $N$ tokens, $x_1, \ldots, x_N$, we encode it with a text encoder to a sequence of vectors
$\hvec_1^m, \ldots, \hvec_N^m \in\RR^d$. 
We then build the span representations using the span start and end token vectors,
$\mvec_{(i,j)} = \text{tanh}((\hvec^m_{i} \oplus \hvec^m_{j})W^a)$,
where $i$ and $j$ are the start and end positions respectively,
$\oplus$ denotes concatenation,
tanh is the activation function,
and $W^a \in\RR^{2d \times d}$ are learnable weights.
For negative instances, we randomly sample spans within the maximum length of 10 from the same input which do not correspond to any entity.
Then we use a learned anchor vector $\svec \in\RR^d$ for contrastive learning, \ie pushing it close to the entity spans and away from negative spans.
\begin{equation}
      L_\textsubscript{pos} = -{\exp(\text{sim}(\svec,  \mvec^+)) \over \sum_{\mvec^\prime\in\mathcal{M}\cup\{\mvec^+\}} \exp(\text{sim}(\svec, \mvec^\prime))},
    \label{eqn:obj_pos}  
\end{equation}
where $\mathcal{M}$ is the negative span set which always contains a special span corresponding to \clstoken, $\mvec^\clstoken=\hvec^m_0$.
However, the above objective alone is not able to determine the prediction of entity spans from null cases at test time. To address this, we further train the model with an extra objective to learn a dynamic threshold using $\mvec^\clstoken$
\begin{equation}
      L_\textsubscript{cls} = -{\exp(\text{sim}(\svec,\mvec^\clstoken) \over \sum_{\mvec^\prime\in\mathcal{M}} \exp(\text{sim}(\svec, \mvec^\prime))}.
    \label{eqn:obj_cls}  
\end{equation}
The overall entity span proposal loss is computed as 
    $L_\textsubscript{span} = (L_\textsubscript{pos} + L_\textsubscript{cls})/2$.
Thus, spans with scores higher than the threshold are predicted as positive.

\noindent \textbf{Entity Linking}
Unlike \citet{wu-etal-2020-scalable} where entity markers are inserted to the entity mention context (the entity mention with surrounding context), we use the raw input sequence as in the entity span proposal task.
For the entity mention context, we pass the input tokens $x_1, \ldots, x_N$ through the entity query encoder to get $\hvec_1^e, \ldots, \hvec_N^e \in\RR^d$.
Then we compute the entity vector based on its start position $i$ and end position $j$, \ie
    $\evec = (\hvec_i^e + \hvec_j^e)/2$.
For entity descriptions, we encode them with the entity description encoder and use the \clstoken~vector $\pvec_e$ as representations. 
The model is trained to match the entity vector with its entity description vector  
\begin{equation}
      L_\textsubscript{link} = -{\exp(\text{sim}(\evec,  \pvec_e^+)) \over \sum_{\pvec^\prime\in\mathcal{P}_e\cup\{\pvec_e^+\}} \exp(\text{sim}(\evec, \pvec^\prime))},
    \label{eqn:obj_link}  
\end{equation}
where $\pvec_e^+$ is the linked description vector and $\mathcal{P}_e$ is the negative entity description set.

\noindent \textbf{Reranking} 
Given a question $Q$ and a passage $P$, we concatenate them as done in expanded query retrieval format $\clstoken~Q~\septoken~P~\septoken$, and encode it using another text encoder.
We use the pair consisting of the \clstoken~vector $\hvec^r_{\clstoken}$ and the first \septoken~vector $\hvec_{\septoken}^r$ from the output for reranking.
The model is trained using the loss
\begin{equation}
\small L_\textsubscript{rank} = -{\exp(\text{sim}(\hvec_{\clstoken}^{r+},  \hvec_{\septoken}^{r+})) \over \sum_{\pvec^{r\prime}\in\mathcal{P}_r\cup\{\pvec^{r+}\}} \exp(\text{sim}(\hvec_{\clstoken}^{r\prime}, \hvec_{\septoken}^{r\prime}))},
    \label{eqn:obj_rank}  
\end{equation}
where $\mathcal{P}_r$ is the set of negative passages concatenated with the same question.
Intuitively, our formulation encourages $\hvec^r_\clstoken$ to capture more information about the question and $\hvec^r_\septoken$ to focus more on the evidence.
The positive pair where the evidence is supportive likely has higher similarity than the negative ones.
Our formulation thus spares the need for an extra task-specific header. 
As the model only learns to rerank single passages, we compute the score for each passage separately for multi-hop cases.  

\subsection{Modular Skill Specialization}
\label{ssec:multi_skill_specialization}
Implementing all aforementioned modules using separate models is apparently inefficient.
As recent work finds that parameter sharing improves the bi-encoder retriever \cite{xiong2020answering}, 
we thus focus on a multi-task learning approach.

One popular choice is to share the text encoder's parameter of all modules \cite{liu-etal-2019-mtdnn}.
However, this approach suffers from task interference, resulting in degraded performance compared with the skill-specific model (\S\ref{ssec:task_interference}).
We attribute the cause to the competition for the model capacity, \ie conflicting signals from different skills require attention to individual syntactic/semantic patterns.
For example, the text encoder for entity-centric queries likely focuses on the local context around the entity 
while the expanded query one tends to represent the latent information based on the relation between the query and previous hop evidence.

Motivated by recent modular approaches for sparse Transformer LM \cite{fedus-et-al-2021-switchtransformer}, we propose to mitigate the task interference by mixing {\it skill-specific Transformer blocks} with shared ones.
A typical Transformer encoder is built with a stack of regular Transformer blocks, each consisting of a multi-head self-attention (MHA) sub-layer and a feed-forward network (FFN) sub-layer,
with residual connections \cite{he2015deep} and layer-normalization \cite{ba2016layer} 
applied to both sub-layers.
The shared Transformer block is identical to a regular Transformer block, \ie all skill inputs are passed through the same MHA and FFN functions.

As shown in \autoref{fig:model_figure}, for skill-specific Transformer blocks,
we select a specialized sub-layer from a pool of $I$ parallel sub-layers based on the input,
\ie different skill inputs are processed independently.
One option is to specialize the FFN expert sub-layer for individual skills, which is widely used by recent mixture-of-expert models \cite{fedus-et-al-2021-switchtransformer,cheng2022taskaware}.
As the FFN sub-layer is found to be important for factual associations \cite{meng2022locating}, we hypothesize that using the popular FFN expert is sub-optimal.
Since most reasoning skills require similar world knowledge, specializing FFN sub-layers likely hinders knowledge sharing.
Instead, different skills typically require the model to attend to distinct input parts.
Thus, we investigate a more parameter-efficient alternative, \ie MHA specialization.
In our experiments, we find it to be more effective in reducing task interference (\S\ref{ssec:task_interference}).

\begin{figure}
    \centering
    \includegraphics[scale=0.5]{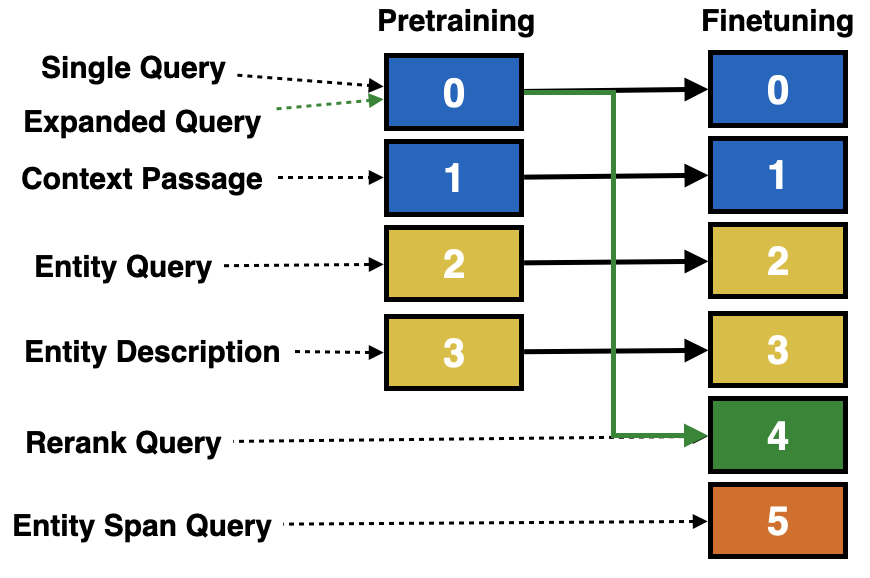}
    \caption{Expert configuration for \ourmodelshort at pretraining and fine-tuning.
    Each numbered box is a skill-specific expert.
    The lines denote input routing where solid ones also indicate weight initialization mappings.
    Green lines highlight the expanded query routing which is different for pretraining and fine-tuning.}
    \label{fig:expert}
\end{figure}

\noindent\textbf{Expert Configuration} Regarding the modularization, a naive setup is to route various task inputs to their dedicated sub-layers (experts), 
\ie two experts for each bi-encoder task (single retrieval, expanded query retrieval and entity linking) and one expert for each cross-encoder task (entity span proposal and reranking), leading to eight experts in total.
To save computation, we make the following adjustments.
Given that single and expanded query retrievers share the same set of target passages, we merge the context expert for both cases.
Due to data sparsity, we find that routing the expanded queries and reranker inputs which are very similar to separate experts is problematic (\S\ref{ssec:task_interference}).
Thus, we merge the expert for expanded queries and reranker inputs. 
During self-supervised pretraining with three bi-encoder tasks, we further share the expert for single and expanded queries for efficiency.
The overall expert configuration is shown in \autoref{fig:expert}.

\noindent\textbf{Multi-task Self-supervision}
Inspired by the recent success of \citet{izacard2021contriever}, we also use \textit{self-supervision} on Wikipedia for pretraining.
Here, we only consider pretraining for bi-encoder skills (\ie single retrieval, expanded query retrieval, and entity linking) where abundant self-supervision is available.
Unlike prior work focusing only on single-type pretraining,
we consider a multi-task setting using individual pages and the hyperlink relations among them.
Specifically, we follow \citet{izacard2021contriever} and \citet{wu-etal-2020-scalable} to construct examples for single retrieval and entity linking, respectively.
For single retrieval, a pair of randomly cropped views of a passage is used as a positive example.
For entity linking, a short text snippet with a hyperlinked entity (entity mention context) is used as the query,
and the first paragraph of its linked Wikipedia page is treated as the target (entity description).
For a given page, we construct an expanded query using a randomly-sampled short text snippet with its first paragraph, and use one first paragraph from linked pages as the target.

\subsection{Inference}
During inference, different skills can be flexibly combined to boost retrieval accuracy.
Those studied configurations are illustrated in \autoref{fig:intro}.
To consolidate the evidence set obtained by different skills, we first align the linking scores based on the same step retrieval scores (single or expanded query retrieval) for sorting.
Documents returned by multiple skills are considered more relevant and thus promoted in ranking.
More details with running examples are provided in \autoref{sec:appendix1}. 

\section{Experiments}
\label{sec:exp}

\begin{table*}[t!]
\centering
\resizebox{\linewidth}{!}{
\begin{tabular}{lcccccccccc}
\toprule
 & \multicolumn{2}{c}{\textbf{NQ}} & \multicolumn{2}{c}{\textbf{WebQ}} & \multicolumn{2}{c}{\textbf{EntityQuestions}} & \multicolumn{2}{c}{\textbf{HotpotQA}}  & \multicolumn{2}{c}{\textbf{Avg}} \\
  &  \textbf{Top-20} & \textbf{Top-100} &  \textbf{Top-20} & \textbf{Top-100}  &  \textbf{Top-20} & \textbf{Top-100} &  \textbf{Top-20} & \textbf{Top-100} &  \textbf{Top-20} & \textbf{Top-100}\\
\midrule
BM25 & 62.9 & 78.3 & 62.4 & 75.5 & \bf 70.8 & \bf 79.2 & 37.5 & 50.5 & 58.4 & 70.9\\
Contriever \cite{izacard2021contriever}  & 67.8 & \bf 82.1 & 65.4 & 79.8 & 61.8 & 74.2  & 48.7 & 64.5 & 60.9 & 75.2  \\
Spider \cite{ram-etal-2022-learning} & \bf 68.3 & 81.2 & 65.9 & 79.7  & 65.1 & 76.4  & 35.3 & 48.6 & 58.7 & 71.5\\
\midrule
\ourmodelshort (pretrain-only) & 68.0 & 81.8 & \bf 66.7 & \bf 80.3 & 70.7 &  79.1 & \bf 77.9 & \bf 87.9 & \bf 70.8 & \bf 82.3 \\
\bottomrule
\end{tabular}
}
\caption{Zero-shot top-$k$ accuracy on test sets for NQ, WebQ and EntityQuestions, and dev set for HotpotQA. 
}
\label{tab:zero-shot}
\end{table*}

\subsection{Datasets}
We consider six popular datasets for evaluation, all focused on Wikipedia,
 with four single-hop data, NQ \cite{kwiatkowski-etal-2019-natural}, WebQ \cite{berant-etal-2013-semantic}, SQuAD \cite{rajpurkar-etal-2016-squad} and EntityQuestions \cite{Sciavolino-etal-2021-simple}; two multi-hop data, HotpotQA \cite{yang-etal-2018-hotpotqa} and OTT-QA \cite{chen2021ottqa}.
 Dataset-specific corpora are used for multi-hop datasets, because HotpotQA requires retrieval hopping between text passages while table-passage hopping is demanded by OTT-QA.
 For single-hop data, we use the Wikipedia corpus from \citet{karpukhin-etal-2020-dense}.
 More detailed (pretraining/fine-tuning) data statistics and experimental settings are in \autoref{sec:appendix2}. 

\subsection{Evaluation Settings}
\label{ssec:eval_setup}
We evaluate our model in three scenarios.

\noindent\textbf{Zero-shot Evaluation}
Similar to recent self-supervised dense retrievers on Wikipedia, we conduct zero-shot evaluations using the retrieval skill from our pretrained model on NQ, WebQ, EntityQuestions and HotpotQA.
To assess the model's ability to handle expanded query retrieval, we design an oracle second-hop retrieval setting (gold first-hop evidence is used) based on HotpotQA.
Following \citet{izacard2021contriever} and \citet{ram-etal-2022-learning}, we report top-$k$ retrieval accuracy (answer recall),
\ie the percentage of questions for which the answer string is found in the top-$k$ passages.

\noindent\textbf{Supervised In-domain Evaluation}
We further fine-tune our pretrained model with two extra skills (entity span proposal and reranking) on NQ, HotpotQA and OTT-QA, again in a multi-task fashion.
Unlike multi-hop data with supervision for all skills, only single retrieval and reranking data is available for NQ.
During training, all datasets are treated equally without any loss balancing.
Different from previous retrieval-only work, we explore \ourmodellong retrieval by using different skill configurations.
Specifically, we use skill configuration for task A, B and C shown in \autoref{fig:intro} for NQ, OTT-QA and HotpotQA, respectively.
We again report top-$k$ retrieval accuracy for NQ and OTT-QA following previous work.
For HotpotQA, we follow the literature using the top-1 pair of evidence accuracy (passage EM).

\noindent\textbf{Cross-data Evaluation}
To test the model robustness towards domain shift,
we conduct cross-data evaluations on SQuAD and EntityQuestions.
Although considerable success has been achieved for supervised dense retrievers using in-domain evaluations,
those models have a hard time generalizing to query distribution shift \citep[\eg questions about rare entities;][]{Sciavolino-etal-2021-simple} compared with BM25.
In particular, we are interested to see whether \ourmodellong retrieval is more robust.
Again, top-$k$ retrieval accuracy is used.

\subsection{Results}
\begin{table}
\centering
\resizebox{\linewidth}{!}{
\begin{tabular}{lcc}
\toprule
 &  \textbf{Top-20} & \textbf{Top-100}  \\
\midrule
DPR-multi \cite{karpukhin-etal-2020-dense} & 79.5 & 86.1\\
ANCE-multi \cite{xiong2021approximate} & 82.1 & 87.9 \\
\midrule
DPR-PAQ \cite{oguz-etal-2022-domain} & 84.7 & 89.2 \\
co-Condenser \cite{gao-callan-2022-unsupervised} & 84.3 & 89.0 \\
SPAR-wiki \cite{chen2021salient} & 83.0 & 88.8 \\
\midrule
\ourmodelshort & \bf 85.6 & \bf 90.2 \\
\bottomrule
\end{tabular}
}
\caption{Supervised top-$k$ accuracy on NQ test.}
\label{tab:sup-nq}
\end{table}

\begin{table}
\centering
\resizebox{\linewidth}{!}{
\begin{tabular}{lccc}
\toprule
 &  \textbf{Top-20} & \textbf{Top-50}  &  \textbf{Top-100}\\
\midrule
CORE \cite{ma2022opendomain} & 74.5 & 82.9 & 87.1 \\
\midrule
\ourmodelshort & 79.9 & \bf 88.9 & \bf 92.2 \\
\ourmodelshort w/ CORE configuration & \bf 80.5 & 88.6 & 91.8 \\ 
\bottomrule
\end{tabular}
}
\caption{Supervised top-$k$ accuracy on OTT-QA dev.
}
\label{tab:sup-ott}
\end{table}

\begin{table}[ht!]
\small
\centering
\resizebox{\linewidth}{!}{
\begin{tabular}{lc}
\toprule
 &  \textbf{Passage EM} \\
\midrule
MDR \cite{xiong2020answering} & 81.20 \\ 
Baleen \cite{khattab2021baleen} & 86.10 \\
\midrule
IRRR \cite{qi-etal-2021-answering} & 84.10 \\ 
TPRR \cite{10.1145/3404835.3462942} & 86.19 \\
\midrule
HopRetriever-plus \cite{hopretriever} & 86.94 \\
AISO \cite{zhu-etal-2021-adaptive} & 88.17 \\
\midrule

\ourmodelshort & \bf 88.89 \\
\bottomrule
\end{tabular}
}
\caption{Supervised passage EM on HotpotQA dev.
}
\label{tab:sup-hotpot}
\end{table}

\noindent\textbf{Zero-shot Results}
\label{ssec:zero_shot}
For zero-shot evaluations, we use two recent self-supervised dense retrievers,
Contriever \cite{izacard2021contriever} and
Spider \cite{ram-etal-2022-learning},
and BM25 as baselines.
The results are presented in \autoref{tab:zero-shot}.
As we can see, BM25 is a strong baseline matching the average retrieval performance of Spider and Contriever over considered datasets.
\ourmodelshort achieves similar results on NQ and WebQ compared with self-supervised dense methods. 
On the other hand, we observe significant gains on HotpotQA and EntityQuestions, where both dense retrievers are lacking.
In summary, our model shows superior zero-shot performance in terms of average answer recall across the board, surpassing BM25 with the largest gains, which indicates the benefit of our multi-task pretraining.

\noindent\textbf{Supervised In-domain Results}
\label{ssec:fine_tune}
As various customized retrievers are developed for NQ, OTT-QA and HotpotQA,
we compare \ourmodelshort with different dataset-specific baselines separately.
For NQ, we report two types of baselines,
1) bi-encoders with multi-dataset training and 2) models with \textit{augmented pretraining}. For the first type, we have DPR-multi \cite{karpukhin-etal-2020-dense} and ANCE-multi \cite{xiong2021approximate}, where the DPR model is initialized from BERT-based and ANCE is initialized from DPR. For the second type, DPR-PAQ \citep{oguz-etal-2022-domain} is initialized from the RoBERTa-large model \cite{liu2019roberta} with pretraining using synthetic queries (the PAQ corpus \cite{lewis-etal-2021-paq}),
co-Condenser \citep{gao-callan-2022-unsupervised} incorporated retrieval-oriented modeling during language model pretraining on Wikipedia; SPAR-wiki \citep{chen2021salient} combine a pretrained lexical model on Wikipedia with a dataset-specific dense retriever. Both co-Condenser and SPAR-wiki are initialized from BERT-base. As shown by results for NQ (\autoref{tab:sup-nq}), \ourmodelshort outperforms all baselines with or without pretraining. It is particularly encouraging that despite being a smaller model, \ourmodelshort achieves superior performance than DPR-PAQ. The reasons are two-fold: \citet{oguz-etal-2022-domain} has shown that scaling up the retriever from base to large size only provides limited gains after pretraining. Moreover, DPR-PAQ only learns a single retrieval skill, whereas \ourmodelshort can combine multiple skills for inference. We defer the analysis of the advantage of chain-of-skills inference later (\S\ref{ssec:cos_inference}).

For OTT-QA, we only compare with the SOTA model CORE \cite{ma2022opendomain}, because other OTT-QA specific retrievers are not directly comparable where extra customized knowledge source is used.
As CORE also uses multiple skills to find evidence chains, we include a baseline where the inference follows the CORE skill configuration but uses modules from \ourmodelshort.
For HotpotQA, we compare against three types of baselines,
dense retrievers focused on expanded query retrieval MDR \cite{xiong2020answering} and Baleen \cite{khattab2021baleen},
sparse retrieval combined with query reformulation IRRR \cite{qi-etal-2021-answering} and TPRR \cite{10.1145/3404835.3462942}
and ensemble of dense, sparse and hyperlink retrieval HopRetriever \cite{hopretriever} and AISO \cite{zhu-etal-2021-adaptive}. The results on OTT-QA and HotpotQA are summarized in \autoref{tab:sup-ott} and \autoref{tab:sup-hotpot}. It is easy to see that \ourmodelshort outperforms all the baselines here, again showing the advantage of our configurable multi-skill model over multiple types of ODQA tasks.
Later, our analyses show that both \ourmodellong inference and pretraining contribute to the observed gains.

\begin{table}
\centering
\resizebox{\linewidth}{!}{
\begin{tabular}{lcccc}
\toprule
 & \multicolumn{2}{c}{\textbf{EntityQuestions}} & \multicolumn{2}{c}{\textbf{SQuAD}} \\
  &  \textbf{Top-20} & \textbf{Top-100} &  \textbf{Top-20} & \textbf{Top-100} \\
\midrule
BM25 & 70.8 & 79.2 & 71.1 & 81.8 \\
DPR-multi \cite{karpukhin-etal-2020-dense}& 56.6 & 70.1 & 52.0 & 67.7 \\
SPAR-wiki \cite{chen2021salient}  & 73.6 & 81.5 & \bf 73.0 & \bf 83.6 \\
\midrule
\ourmodelshort & \bf 76.3 & \bf 82.4 & 72.6 & 81.2 \\
\bottomrule
\end{tabular}
}
\caption{Cross-dataset top-$k$ accuracy on test sets.}
\label{tab:cross_eval}
\end{table}
\noindent\textbf{Cross-data Results}\label{ssec:cross_eval}
Given that both EntityQuestions and SQuAD are single-hop, we use baselines on NQ with improved robustness for comparison.
Particularly, SPAR-wiki is an ensemble of two dense models with one pretrained using BM25 supervision on Wikipedia and the other fine-tuned on NQ.
BM25 is included here, as it is found to achieve better performance than its dense counterpart on those two datasets.
The evaluation results are shown in \autoref{tab:cross_eval}.
Overall, our model achieves the largest gains over BM25 on both datasets,
indicating that our multi-task fine-tuned model with \ourmodellong inference is more robust than previous retrieval-only approaches.

\section{Analysis}
\label{sec:analysis}

\begin{table}
\centering
\resizebox{\linewidth}{!}{
\begin{tabular}{lccc}
\toprule
 & \#Params & \textbf{Top-20}  &  \textbf{Top-100}\\
\midrule
\ourmodellong inference\\
\midrule
No Expert & 111M & 90.2 & 92.4 \\
FFN Expert(naive) & 252M  & 91.3 & 93.4 \\
MHA Expert(naive) & 182M  & 92.0 & 94.0 \\
MHA Expert(COS) & 182M & 92.0 & 94.2 \\
\midrule
Retrieval-only inference \\
\midrule
Multi-hop Retriever & 110M & 85.1 & 88.9 \\
MHA Expert(naive) & 182M  & 82.8 & 87.0 \\
MHA Expert(COS) & 182M & 85.9 & 89.6 \\
\bottomrule
\end{tabular}
}
\caption{Ablation results on HotpotQA dev using top-$k$ retrieval accuracy. All models are initialized from BERT-base and trained on HotpotQA only.}
\label{tab:abla-hotpot}
\end{table}

\subsection{Task Interference}
\label{ssec:task_interference}
We conduct ablation studies on HotpotQA to compare different ways of implementing skill-specific specialization (discussed in \S\ref{ssec:multi_skill_specialization}) and their effects on task interference.
As MHA experts are used for our model, we consider two variants for comparison: 1) the no-expert model where all tasks share one encoder, and 2) the FFN expert model where specialized FFN sub-layers are used.
Then we also compare the proposed expert configuration with a variant where the expanded query retrieval inputs share the same expert as single retrieval, denoted as the naive setting.
The results are shown in the upper half of \autoref{tab:abla-hotpot}.
Compared with the no-expert model, both FFN and MHA experts can effectively reduce task interference, wherein MHA expert is more effective overall.
Our proposed expert configuration can further help.
\begin{figure}
    \centering
    \includegraphics[scale=0.23]{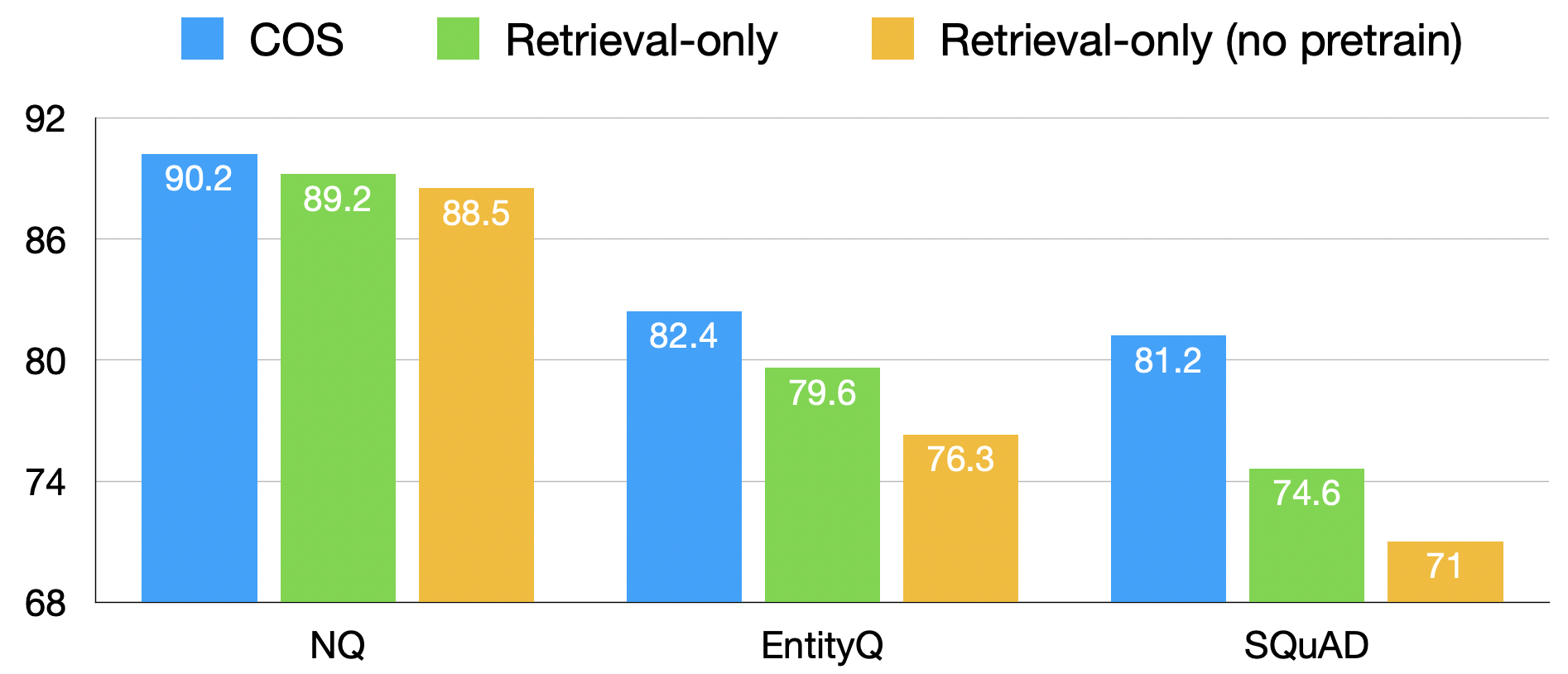}
    \caption{Top-$100$ retrieval accuracy on inference strategy: \ourmodellong vs retrieval-only.}
    \label{fig:skills}
\end{figure}

\begin{figure}
    \centering
    \includegraphics[scale=0.25]{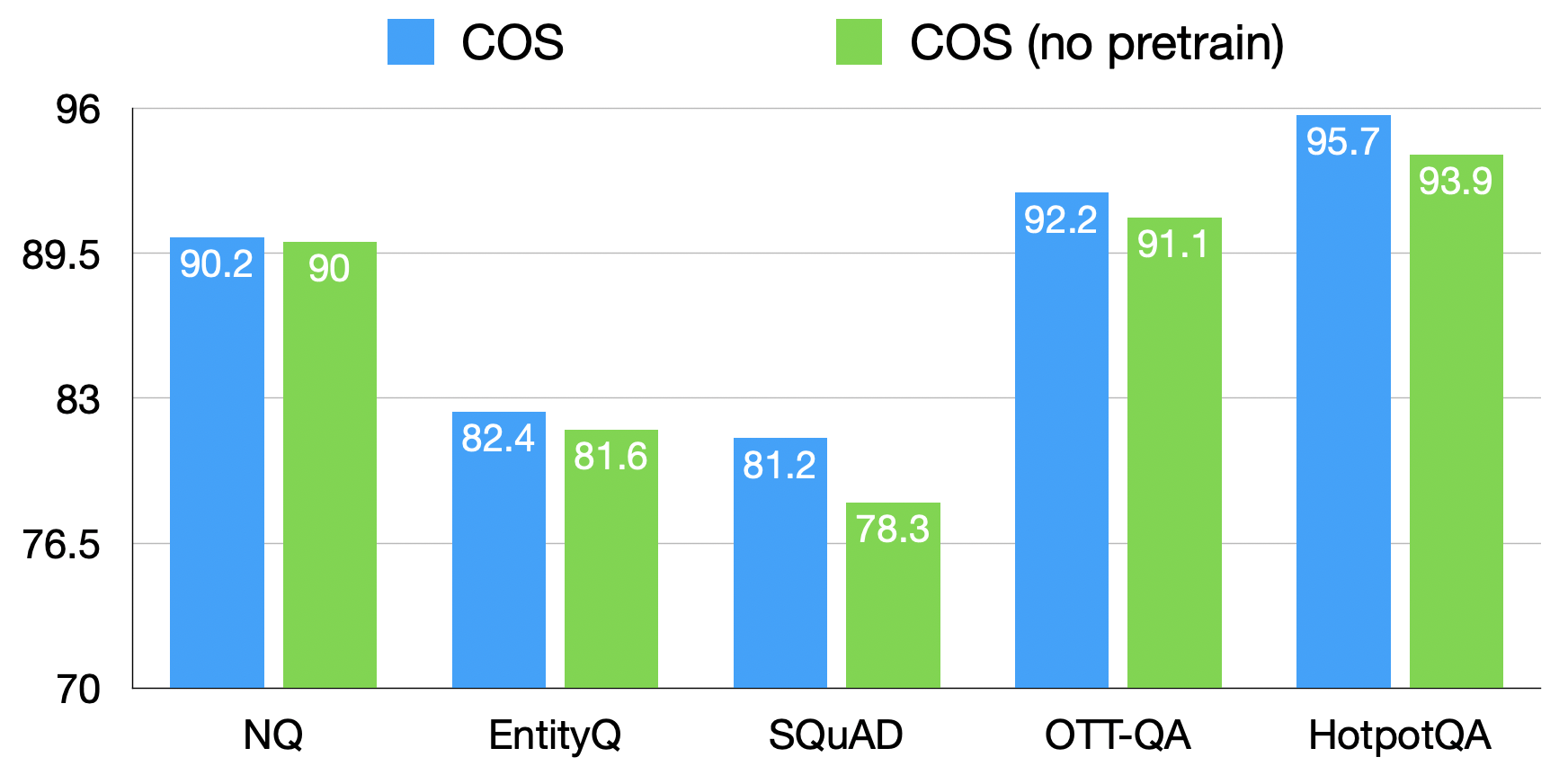}
    \caption{Comparison on the effect of pretraining using top-$100$ retrieval accuracy with COS inference.
    }
    \label{fig:pretrain}
\end{figure}

\subsection{Benefit of \ourmodellong Inference}
\label{ssec:cos_inference}
Here we explore the benefits of the chained skill inference over the retrieval-only version. We additionally train a multi-hop retriever following \citet{xiong2020answering}, and compare it with the two MHA expert models using the same two rounds of retrieval-only inference.
The comparison is shown in the lower part of \autoref{tab:abla-hotpot}.
As we can see, retrieval-only inference suffers large drops in performance.
Although our proposed and naive MHA expert configurations have similar performance using \ourmodellong inference, the naive configuration model shows severe degradation caused by task interference compared with the multi-hop retriever,
validating the effectiveness of our proposed model.
We further compare our \ourmodellong inference with the retrieval-only inference on NQ, EntityQuestions and SQuAD in \autoref{fig:skills}.
It is easy to see that our pretraining can benefit the retrieval-only version.
However, using better skill configurations via \ourmodellong inference yields further improvements, particularly on those unseen datasets.

\subsection{Effect of Pretraining}
\label{ssec:pretraining}
To further demonstrate the benefit of our proposed multi-task pretraining, we fine-tune another multi-task model following the same training protocol as \ourmodelshort but BERT model weights are used for initialization.
Both \ourmodelshort and the model without pretraining are then using the same skill configuration for inference.
The results are illustrated in \autoref{fig:pretrain}.
Similar to the retrieval-only version (\autoref{fig:skills}), we find that \ourmodelshort consistently outperforms the multi-task model without pretraining across all considered datasets using \ourmodellong inference.
Again, the pretrained model is found to achieve improvements across the board, especially on out-of-domain datasets, which validates the benefits of our multi-task pretraining.

\begin{table}
\centering
\resizebox{\linewidth}{!}{
\begin{tabular}{lcccc}
\toprule
 & Query & Doc & \textbf{Top-20}  &  \textbf{Top-100}\\
\midrule
Single query* & 0 & 1 & \bf 96.1 & \bf 98.2 \\
Single query & 4 & 1 & 90.1 & 95.2 \\
Single query & 2 & 1 & 91.8 & 95.9 \\
Single query & 2 & 3 & 87.4 & 92.7\\
\midrule
Expanded query & 0 & 1 & 94.2 & 97.0 \\
Expanded query* & 4 & 1 & \bf 95.3 & \bf 97.4 \\
Expanded query & 2 & 1 & 74.5 & 85.8\\
Expanded query & 2 & 3 & 67.3 & 79.6 \\
\bottomrule
\end{tabular}
}
\caption{Results of feeding the inputs to different experts, where the first two columns represent the query expert id and document expert id. * denotes the proposed setup}
\label{tab:swap-expert}
\end{table}
\subsection{Swapping Experts}
To understand if different experts in our model learned different specialized knowledge, we experiment with swapping experts for different inputs on HotpotQA. In particular, we feed the single query input and expanded query input to different query experts and then retrieve from either the context passage index or the entity description index. For single query input, we measure if the model can retrieve one of the positive passages. For expanded query input, we compute the recall for the other positive passage as done in (\S\ref{ssec:zero_shot}).
The results are shown in \autoref{tab:swap-expert}. Although both the single query expert and the expanded query expert learn to retrieve evidence using the \clstoken token, swapping the expert for either of these input types leads to a significant decrease in performance. Also, switching to the entity query expert and retrieving from the entity description index results in a large drop for both types of inputs. This implies that each specialized expert acquires distinct knowledge and cannot be substituted for one another.

\section{Question Answering Experiments}
Here, we conduct end-to-end question-answering experiments on NQ, OTT-QA and HotpotQA, using retrieval results from \ourmodelshort.
Following the literature, we report exact match (EM) accuracy and F1 score.

For NQ and OTT-QA, we re-implement the Fusion-in-Encoder (FiE) model \cite{kedia2022fie} because of its superior performance on NQ.
For NQ, the model reads top-100 passages returned by \ourmodelshort, and for OTT-QA, the model reads top-50 evidence chains, in order to be comparable with previous work.
Here, separate models are trained for each dataset independently.
Due to space constraints, we only present the results on OTT-QA and leave the NQ results to \autoref{tab:nq-qa}.
The OTT-QA results are summarized in \autoref{tab:ott-qa}.
Our model, when coupled with the FiE, is able to outperform the previous baselines by large margins on OTT-QA, and we can see that the superior performance of our model is mainly due to \ourmodelshort.

Finally, for HotpotQA, since the task requires the model to predict supporting sentences in addition to the answer span, 
we follow \citet{zhu-etal-2021-adaptive} to train a separate reader model to learn answer prediction and supporting sentence prediction jointly.
Due to space constraints, we leave the full results to \autoref{tab:hotpot-qa}.
Overall, our method achieves competitive QA performance against the previous SOTA with improved exact match accuracy. 

\begin{table}
\centering
\resizebox{\linewidth}{!}{
\begin{tabular}{lcccc}
\toprule
 & \multicolumn{2}{c}{Dev} & \multicolumn{2}{c}{Test} \\
 & \textbf{EM} &  \textbf{F1} & \textbf{EM} &  \textbf{F1}\\
\midrule
HYBRIDER \cite{chen-etal-2020-hybridqa} & 10.3 & 13.0 & 9.7 & 12.8 \\
FR+CBR\cite{chen2021ottqa}  & 28.1 & 32.5 & 27.2 & 31.5 \\
CARP \cite{zhong2022reasoning} & 33.2 & 38.6 & 32.5 & 38.5 \\
OTTer \cite{huang2022mixedmodality} & 37.1 & 42.8 & 37.3 & 43.1 \\
CORE \cite{ma2022opendomain} & 49.0 & 55.7 & 47.3 & 54.1 \\
\midrule
CORE + FiE & 51.4  & 57.8 & - & - \\
\ourmodelshort + FiE & \bf 56.9 & \bf 63.2 & \bf 54.9 & \bf 61.5 \\
\bottomrule
\end{tabular}
}
\caption{End-to-end QA results on OTT-QA.}
\label{tab:ott-qa}
\end{table}

\section{Related Work}
Dense retrievers are widely used in recent literature for ODQA \cite{lee-etal-2019-latent,karpukhin-etal-2020-dense}. While most previous work focuses on single retrieval \cite{xiong2021approximate,qu-etal-2021-rocketqa}, some efforts have also been made towards better handling of other query types. \citet{xiong2020answering} propose a joint model to handle both single retrieval and expanded query retrieval.
\citet{chen2021salient} train a dense model to learn salient phrase retrieval.  
\citet{ma2022opendomain} build an entity linker to handle multi-hop retrieval.
Nevertheless, all those models are still customized for specific datasets,
\eg only a subset of query types are considered
or separate models are used,
making them un-reusable and computationally intensive.
We address these problems by pinning down a set of functional skills that enable joint learning over multiple datasets. 

Mixure-of-expert models have also become popular recently \cite{fedus-et-al-2021-switchtransformer}. Methods like gated routing \cite{lepikhin2020gshard} or stochastic routing of experts \cite{zuo2021taming} do not differentiate the knowledge learned by different experts. 
Instead, our work builds expert modules that learn reusable skills which can be flexibly combined for different use cases. 

Another line of work focus on unsupervised dense retrievers using self-supervised data constructed from the inverse-cloze-task \cite{lee-etal-2019-latent}, random croppings \cite{izacard2021contriever}, truncation of passages with the same span \cite{ram-etal-2022-learning}, hyperlink-induced passages \cite{zhou-etal-2022-hyperlink} or synthetic QA pairs \cite{oguz-etal-2022-domain}.
Other model architecture adjustments on Transformer for retrieval are proposed \cite{gao-callan-2021-condenser,gao-callan-2022-unsupervised}.
Our work can be viewed as a synergy of both.
Our multi-task pretrained model can perform better zero-shot retrieval.
Our modular retriever can be further fine-tuned in a multi-task fashion to achieve better performance.

\section{Conclusions}
In this work, we propose a modular model \ourmodellong (\ourmodelshort) that learns five reusable skills for ODQA via multi-task learning.
To reduce task interference, we design a new parameterization for skill modules.
We also show that skills learned by \ourmodelshort can be flexibly chained together to better fit the target task.
\ourmodelshort can directly perform superior zero-shot retrieval using multi-task self-supervision on Wikipedia.
When fine-tuned on multiple datasets, \ourmodelshort achieves SOTA results across the board.
For future work, we are interested in exploring scaling up our method and other scenarios, \eg commonsense reasoning \cite{talmor2022commonsenseqa} and biomedical retrieval \cite{2020bioasq,zhang-etal-2022-knowledge}. 

\section*{Acknowledgements}
We would like to thank Aman Madaan, Sheng Zhang, and other members of the Deep Learning
group at Microsoft Research for their helpful discussions and anonymous reviewers for their valuable suggestions on this paper.

\section*{Limitations}
We identify the following limitations of our work.

Our current \ourmodelshort's reranking expert only learns to rerank single-step results.
Thus it can not model the interaction between documents in case of multi-passage evidence chains, which might lead to sub-optimal performance, \eg when we need to rerank the full evidence path for HotpotQA.
At the same time, we hypothesize that the capacity of the small model used in our experiments is insufficient for modeling evidence chain reranking. We leave the exploration of learning a full path reranker for future work. 

Also, our current pretraining setup only includes the three bi-encoder tasks,
and thus we can not use the pretrained model out-of-box to solve tasks like end-to-end entity linking.
Consequently, the learned skills from self-supervision can not be chained together to perform configurable zero-shot retrieval.
It would be interesting to also include the entity span proposal skill in the pretraining stage, which could unleash the full potential of the \ourmodellong inference for zero-shot scenarios.

\bibliography{anthology,custom}
\bibliographystyle{acl_natbib}

\clearpage
\appendix

\setcounter{table}{0}
\renewcommand{\thetable}{A\arabic{table}}

\setcounter{figure}{0}
\renewcommand{\thefigure}{A\arabic{figure}}

\begin{figure*}
    \centering
    \includegraphics[scale=0.36]{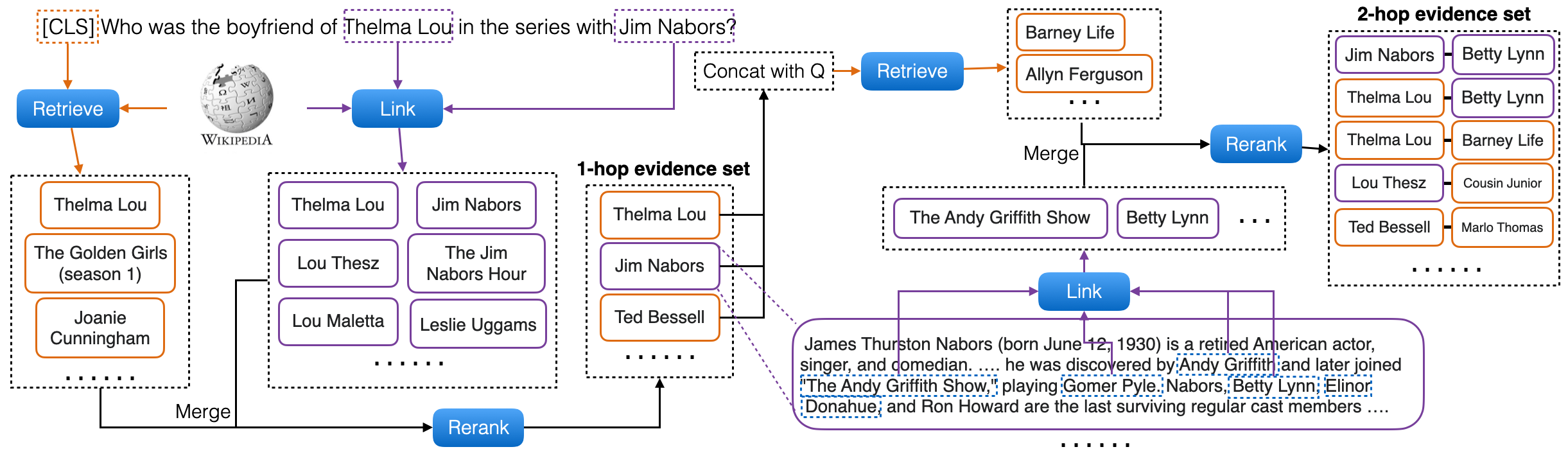}
    \caption{The reasoning pipeline of \ourmodelfull. Given a question, \ourmodelshort first identifies salient spans in the question, then the retrieving and linking skills are both used to find first-hop evidence, using the \clstoken token and entity mention representation respectively. Then we merge all the evidence through score alignment and the reranking skill. For top-ranked evidence documents, we concatenate each of them with the question and perform another round of retrieving and linking. Then the second hop evidence are merged and reranked in the same fashion. Finally, the reasoning paths are sorted based on both hops' scores}
    \label{fig:pipeline}
\end{figure*}
\section{Inference Pipeline}
\label{sec:appendix1}

At inference time, our model utilizes the retrieving skill or the linking skill or both in parallel to gather evidence at every reasoning step. When both skills are used, one problem is that the scores associated with the evidence found by different skills are not aligned, \ie naively sorting the retrieved documents and linked documents together may cause one pool of documents to dominate over the other. Thus we propose to align the linking scores based on the same step retrieval score:
\begin{equation}
    \text{ls}_i = \text{ls}_i / \text{max}(\{\text{ls}\}\cup\{\text{rs}\}) \times \text{max}(\{\text{rs}\}),
\end{equation}
where $\text{ls}_i$ represents the linking score of the document $i$ and $\{\text{ls}\}$, $\{\text{rs}\}$ represent the set of linking scores and retrieving scores for top-$K$ documents from each skill. Effectively, if the raw linking score is larger than the retrieving score, we would align the top-1 document from each set. On the other hand, if the raw linking score is smaller, it would not get scaled. The reason is that certain common entities may also be detected and linked by our model \eg United States, but they usually do not contribute to the answer reasoning, thus we do not want to encourage their presence. 

In the case of a document being discovered by both skills, we promote its ranking in the final list. To do so, we take the max of the individual score (after alignment) and then multiply by a coefficient $\alpha$, which is a hyper-parameter. 

\begin{equation}
    s_i = \alpha \; \text{max}(\text{ls}_i, \text{rs}_i). 
\end{equation}
Finally, we use the reranking skill to compute a new set of scores for the merged evidence set, and then sort the documents using the combination of retrieving/linking score and reranking score: 
\begin{equation}
    s_i + \beta \; \text{rankscore}_i.
\end{equation}
$\beta$ is another hyper-parameter. For multi-hop questions, the same scoring process is conducted for the second-hop evidence documents and then the two-hop scores are aggregated to sort the reasoning chains. 
The inference pipeline is also illustrated in \autoref{fig:pipeline}. 

\section{Experimental Details}
\label{sec:appendix2}
\subsection{Data Statistics}
The detailed data statistics are shown in \autoref{tab:stats}. 

\noindent \textbf{Pretraining} 
We follow \citet{izacard2021contriever} and \citet{wu-etal-2020-scalable} to construct examples for single retrieval and entity linking, respectively.
For single retrieval, a pair of randomly cropped views of a passage is treated as a positive example.
Similar to Spider \cite{ram-etal-2022-learning}, we also use the processed DPR passage corpus based on the English Wikipedia dump from 2018/12/20.
For entity linking, we directly use the preprocessed data released by BLINK \cite{wu-etal-2020-scalable} based on the English Wikipedia dump from 2019/08/01.
For expanded query retrieval, we construct the pseudo query using a short text snippet with the first passage from the same page, and we treat the first passage from linked pages as the target.
As no hyperlink information is preserved for the DPR passage corpus, we use the English Wikipedia dump from 2022/06/01 for data construction. In each Wikipedia page, we randomly sample 30 passages with hyperlinks. (If there are less than 30 passages with hyperlinks, we take all of them.) Each sampled passage, together with the first passage of the page, form a pseudo query. Then, in each sampled passage, we randomly pick an anchor entity and take the first passage of its associated Wikipedia page as the target. To avoid redundancy, if an anchor entity has been used 10 times in a source page, we no longer pick it for the given source. If the query and the target together exceed 512 tokens, we will truncate the longer of the two by randomly dropping its first token or its last token.

\noindent \textbf{Finetuning} For NQ, we adopted the retriever training data released by \citet{ma-etal-2022-open} and further used them for the reranking skill. Note that data from \citet{ma-etal-2022-open} also contains table-answerable questions in NQ, and we simply merged the corresponding training splits with the text-based training split. That's why the number of examples in the last column is greater than the number of questions in the training set. 

For HotpotQA, we adopted single retrieval and expanded query retrieval data released by \citet{xiong2020answering}. For question entity linking data, we heuristically matched the entity spans in the question with the gold passages' title to construct positive pairs, and we use the same set of negative passages as in single retrieval. For passage entity linking, we collected all unique gold passages in the training set and their corresponding hyperlinks for building positives and mined negatives using BM25. Finally, the reranking data is the same as single retrieval. 

For OTT-QA, we adopt the single retrieval and table entity linking data released by \citet{ma2022opendomain}. For expanded query retrieval, we concatenate the question with the table title, header, and row that links to the answer-containing passage as the query, and the corresponding passage is treated as a positive target. The negatives are mined with BM25. Finally, reranking data is the same copy as in single retrieval except that we further break down tables into rows and train the model to rank rows. This is because we want to make the reranking and expanded query retrieval more compatible.

Since iterative training is shown to be an effective strategy by previous works \cite{xiong2021approximate,ma-etal-2022-open}, we further mined harder negatives for HotpotQA and OTT-QA skill training data. Specifically, we train models using the same configuration as in pretraining (four task-specific experts, with no reranking data or span proposal data) for HotpotQA and OTT-QA respectively (models are initialized from BERT-based-uncased). Then we minded harder negatives for each of the data types using the converged model. The reranking and the entity span proposal skills are excluded in this round because the reranking can already benefit from harder negative for single retrieval (as two skills share the same data) and the entity span proposal does not need to search through a large index. Finally, the data splits coupled with harder negatives are used to train our main \ourmodelfull and conduct ablation studies.

\begin{table*}
\centering
\begin{tabular}{lccclc}
\toprule
Dataset & Train & Dev & Test & Skill Training Data & \# Examples \\
\midrule
\multirow{3}{*}{Pretraining} & 
\multirow{3}{*}{-} & \multirow{3}{*}{-} & \multirow{3}{*}{-} & single retrieval & 6M \\ 
 & & & & expanded query retrieval & 6M \\
 & & & & passage entity linking & 9M \\
\midrule
\multirow{2}{*}{NQ} & \multirow{2}{*}{79,168}  & \multirow{2}{*}{8,757} & \multirow{2}{*}{3,610} & single retrieval & 86,252 \\ 
 & & & & reranking & 86,252 \\
 \midrule
\multirow{5}{*}{HotpotQA} & \multirow{5}{*}{90,447} & \multirow{5}{*}{7,405} & \multirow{5}{*}{7,405} & single retrieval & 90,447 \\
& & & & expanded query retrieval & 90,447 \\
& & & & question entity linking & 80,872  \\
& & & & passage entity linking & 104,335 \\
& & & & reranking & 90,447 \\
\midrule
\multirow{4}{*}{OTT-QA} & \multirow{4}{*}{41,469} & \multirow{4}{*}{2,214} & \multirow{4}{*}{2,158} & single retrieval & 41,469 \\ 
& & & & expanded query retrieval & 31,638 \\
& & & & table entity linking & 19,764 \\
& & & & reranking & 41,479 \\
\midrule
EntityQuestions & - & 22,068 & 22,075 & - & -  \\
WebQ & - & - & 2,032 & - & -  \\
SQuAD & - & - & 10,570 & - & -  \\

\bottomrule
\end{tabular}
\caption{Statistics of datasets used in our experiments, columns 2-4 represent the number of questions in each split. The last two columns contain the type of training data and the corresponding number of instances}
\label{tab:stats}
\end{table*}

\subsection{Training Details}
\noindent \textbf{Pretraining}
Similar to Contriever \cite{izacard2021contriever},
we adopt a continual pretraining setup based on the uncased BERT-base architecture, but our model is initialized from the Contriever weights. 
We train the model for 20 epochs with the batch size of 1024 and the max sequence length of 256.
Here, we only use in-batch negatives for contrastive learning.
The model is optimized using Adam with the initial learning rate of 1e-4.
The final checkpoint is used for fine-tuning later.

\noindent \textbf{Finetuning} When initializing from pretrained \ourmodelshort, the weights mapping for the first 5 experts are illustrated in \autoref{fig:expert} and the last expert is initialized from BERT-base-uncased. For all experiments, we train models for 40 epochs with the batch size of 192, the learning rate of 2e-5, and the max sequence length of 256. During training, each batch only contains training data for one of the skills from one dataset, thus the model can effectively benefit from the in-batch negatives. To train the entity span proposal skill, we use the same data as entity linking. In particular, we route the data to span proposal experts 20\% of the time otherwise the data go through entity linking experts.

\subsection{Inference Details}
\noindent \textbf{Zero-shot-evaluation} We directly use the single retrieval skill to find the top100 documents and compute the results in \autoref{tab:zero-shot}.

\noindent \textbf{Supervised and Cross-dataset}
For NQ, EntityQuestions and SQuAD, the reasoning path has a length of 1, \ie only single passages. We use both single retrieval and linking skills to find a total of top 1000 passages first, and then reduce the set to top 100 using the reranking skill.

Both HotpotQA and OTT-QA have reasoning paths with max length 2. 
For OTT-QA, we first find top 100 tables using the single retrieval skill following \cite{ma2022opendomain}. Then we break down tables into rows and use the reranking skill to keep only top 200 rows. Then for each row, expanded query retrieval and linking skills are used to find the second-hop passages, where we keep top 10 passages from every expanded query retrieval and top 1 passage from every linked entity. Finally, we apply the same heuristics, as done in \citet{ma2022opendomain} to construct the final top 100 evidence chains. 

For HotpotQA, single retrieval and linking are used jointly to find the first-hop passages where we keep top 200 passages from single retrieval and top 5 passage from each linked question entity. The combined set is then reranked to keep the top 30 first-hop passages. Then expanded query retrieval and passage entity linking are applied to these 30 passages, where we keep top 50 passages from expanded query retrieval and top 2 passages from every linked passage entity. 
Next, another round of reranking is performed on the newly collected passages and then we sort the evidence passage chains based on the final aggregated score and keep top 100 chains. Since all of the baselines on HotpotQA adopt a large passage path reranker, we also trained such a model following \cite{zhu-etal-2021-adaptive} (discussed in Appendix \ref{sec:appendix4}) to rank the top 100 passage chains to get the top 1 prediction. 

The hyperparameters for OTT-QA and HotpotQA inference are selected such that the total number of evidence chains are comparable to previous works \cite{ma2022opendomain,xiong2020answering}. 


\begin{table}
\centering
\resizebox{\linewidth}{!}{
\begin{tabular}{lcc}
\toprule
 & \#Params & \textbf{EM} \\
\midrule
FiD \cite{izacard-grave-2021-leveraging} & 770M & 51.4 \\
UnitedQA-E \cite{cheng-etal-2021-unitedqa} & 330M & 51.8 \\
FiD-KD \cite{izacard2020distilling} & 770M & 54.4 \\
EMDR$^2$ \cite{NEURIPS2021_da3fde15} & 440M & 52.5 \\ 
YONO \cite{lee2021need} & 440M & 53.2 \\
UnitedQA \cite{cheng-etal-2021-unitedqa} & 1.87B & 54.7 \\
R2-D2 \cite{fajcik-etal-2021-r2-d2} & 1.29B & 55.9 \\
FiE \cite{kedia2022fie} & 330M & \bf 58.4 \\
\midrule
FiE (ours implementation) & 330M & 56.3 \\
\ourmodelshort + FiE & 330M & 56.4 \\
\bottomrule
\end{tabular}
}
\caption{End-to-end QA Exact Match score on NQ}
\label{tab:nq-qa}
\end{table}

\section{Question Answering Results}
\label{sec:appendix4}
\subsection{Training Details}
We follow descriptions in \cite{kedia2022fie} for re-implementation of FiE model and the model is initialized from Electra-large \cite{clark2020electra}. For NQ, we train the model for 5,000 steps with the effective batch size of 64, the learning rate of 5e-5, the layer-wise learning rate decay of 0.9, the max answer length of 15, the max question length of 28, the max sequence length of 250, and 10 global tokens. Note that although \citet{kedia2022fie} reports that training with 15,000 steps leads to better performance, we actually found it to be the same as 5,000 steps. Thus we train with fewer steps to save computation. For OTT-QA, we used the same set-up of hyperparameters except that the max sequence length is changed to 500. 

\begin{table*}
\centering
\resizebox{\linewidth}{!}{
\begin{tabular}{@{\extracolsep{4pt}}lcccccccccccc}
\toprule
 & \multicolumn{6}{c}{Dev} & \multicolumn{6}{c}{Test} \\\cline{2-7}\cline{8-13}
  & \multicolumn{2}{c}{Ans} & \multicolumn{2}{c}{Sup} & \multicolumn{2}{c}{Joint} & \multicolumn{2}{c}{Ans} & \multicolumn{2}{c}{Sup} & \multicolumn{2}{c}{Joint} \\ \cline{2-3}\cline{4-5}\cline{6-7}\cline{8-9}\cline{10-11}\cline{12-13}
 & EM &  F1 & EM &  F1 & EM &  F1 & EM &  F1 & EM &  F1 & EM &  F1  \\
\midrule
MUPPET \cite{feldman-el-yaniv-2019-multi} & 31.1 & 40.4 & 17.0 & 47.7 & 11.8 & 27.6 & 30.6 & 40.3 & 16.7 & 47.3 & 10.9 & 27.0 \\
CogQA \cite{ding-etal-2019-cognitive} & 37.6 & 49.4 & 23.1 & 58.5 & 12.2 & 35.3 & 37.1 & 48.9 & 22.8 & 57.7 & 12.4 & 34.9 \\
GoldEn Retriever \cite{qi-etal-2019-answering} & - & - & - & - & - & - & 37.9 & 49.8 & 30.7 & 64.6 & 18.0 & 39.1 \\
Semantic Retrieval \cite{nie-etal-2019-revealing} & 46.5 & 58.8 & 39.9 & 71.5 & 26.6 & 49.2 & 45.3 & 57.3 & 38.7 & 70.8 & 25.1 & 47.6 \\
Transformer-XH \cite{Zhao2020Transformer-XH:} & 54.0 & 66.2 & 41.7 & 72.1 & 27.7 & 52.9 & 51.6 & 64.1 & 40.9 & 71.4 & 26.1 & 51.3 \\
HGN \cite{fang-etal-2020-hierarchical} & - & - & - & - & - & - & 59.7 & 71.4 & 51.0 & 77.4 & 37.9 & 62.3 \\
GRR \cite{asai2020learning} & 60.5 & 73.3 & 49.2 & 76.1 & 35.8 & 61.4 & 60.0 & 73.0 & 49.1 & 76.4 & 35.4 & 61.2 \\
DDRQA \cite{ddrqa} & 62.9 & 76.9 & 51.3 & 79.1 & - & - & 62.5 & 75.9 & 51.0 & 78.9 & 36.0 & 63.9 \\
MDR \cite{xiong2020answering} & 62.3 & 75.1 & 56.5 & 79.4 & 42.1 & 66.3 & 62.3 & 75.3 & 57.5 & 80.9 & 41.8 & 66.6 \\
IRRR+ \cite{qi-etal-2021-answering} & - & - & - & - & - & - & 66.3 & 79.9 & 57.2 & 82.6 & 43.1 & 69.8 \\
HopRetriever-plus \cite{hopretriever} & 66.6 & 79.2 & 56.0 & 81.8 & 42.0 & 69.0 & 64.8 & 77.8 & 56.1 & 81.8 & 41.0 & 67.8 \\
TPRR \cite{10.1145/3404835.3462942} & 67.3 & 80.1 & 60.2 & 84.5 & 45.3 & 71.4 & 67.0 & 79.5 & 59.4 & 84.3 & 44.4 & 70.8 \\
AISO \cite{zhu-etal-2021-adaptive}& 68.1 & 80.9 & \bf 61.5 & \bf 86.5 & 45.9 & \bf 72.5 & \bf 67.5 & \bf 80.5 & 61.2 & \bf 86.0 & 44.9 & \bf 72.0 \\
\midrule
\ourmodelshort & \bf 68.2 & \bf 81.0 & 61.1 & 85.3 & \bf 46.4 & 72.3 & 67.4 & 80.1 & \bf 61.3 & 85.3 & \bf 45.7 & 71.7   \\
\bottomrule
\end{tabular}
}
\caption{End-to-end QA results on Hotpot-QA.}
\label{tab:hotpot-qa}
\end{table*}

For HotpotQA path reranker and reader, we prepare the input sequence as follows: "\clstoken Q \septoken yes no \texttt{[P]} P1 \texttt{[P]} P2 \septoken", where \texttt{[P]} is a special token to denotes the start of a passage. Then the input sequence is encoded by the model and we extract passage start tokens representations $p_1, ... p_m$ and averaged sentence embeddings for every sentence in the input $s_1, ... s_n$ to represent passages and sentences respectively. The path reranker is trained with three objectives: passage ranking, supporting sentence prediction and answer span extraction, as we found the latter two objectives also aid the passage ranking training. For answer extraction, the model is trained to predict the start and end token indices as commonly done in recent literature \cite{xiong2020answering,zhu-etal-2021-adaptive}. For both passage ranking and supporting sentence prediction, the model is trained with the ListMLE loss \cite{10.1145/1390156.1390306}. In particular, every positive passage in the sequence is assigned a label of 1, and every negative passage is assigned 0. To learn a dynamic threshold, we also use the \clstoken token $p_0$ to represent a pseudo passage and assign a label of 0.5. Finally, the loss is computed as follows:  
\begin{equation}
      L_\textsubscript{p} = - \sum_{i=0}^m \log {\exp(p_i W_p) \over \sum_{p^\prime\in\mathcal{P}\cup\{p_i\}} \exp(p^\prime W_p)}.
    \label{eqn:obj_listmle}  
\end{equation}
where $\mathcal{P}$ contains all passages representations that have labels smaller than $p_i$. $W_p \in \mathrm{R}^{d}$ are learnable weights and d is the hidden size. In other words, the model learns to assign scores such that positive passages $>$ thresholds $>$ negative passages. The supporting sentence prediction is also trained using \autoref{eqn:obj_listmle}. Overall, use the following loss weighting:
\begin{equation}
      L_\textsubscript{path} = L_p + L_a + 0.5 \times L_s
    \label{eqn:3losses}  
\end{equation}
where $L_a$ is the answer extraction loss and $L_s$ is the supporting sentence prediction loss. 

During training, we sample 0-2 positive passages and 0-2 negative passages from the top 100 chains returned by \ourmodelshort, and the model encodes at most 3 passages, \ie the passage chain structure is not preserved and the passages are sampled independently. We train the model for 20,000 steps with the batch size of 128, the learning rate of 5e-5, the layer-wise learning rate decay of 0.9, the max answer length of 30, the max question length of 64, and the max sequence length of 512. For inference, the model ranks top 100 passage chains with structure preserved. We sum the scores of the two passages in every chain and subtract the dynamic threshold score and sort the chains based on this final score. 

Next, we train a reader model that only learns answer extraction and supporting sentence prediction. We only train the model using the two gold passages with the following loss weighting. 
\begin{equation}
      L_\textsubscript{reader} = L_a + 0.5 \times L_s
    \label{eqn:2losses}  
\end{equation}
The model uses the same set of hyperparameters as the path reranker except that the batch size is reduced to 32. At inference time, the model directly read the top 1 prediction returned by the path reranker. Both models here are initialized from Electra-large.

\subsection{Results}
The NQ results are presented in \autoref{tab:nq-qa}. Overall, our model achieves a similar performance as our own FiE baseline. FiE baseline uses the reader data released by the FiD-KD model, which has an R100 of 89.3 (vs 90.2 of \ourmodelshort). Considering that the gap between our method and FiD-KD model's top 100 retrieval recall is relatively small, this result is not surprising.  

The HotpotQA results are shown in \autoref{tab:hotpot-qa}. Overall our results are similar to previous SOTA methods on the dev set. At the time of the paper submission, we have not got the test set results on the leaderboard. 

We adopted DPR evaluation scripts \footnote{\url{https://github.com/facebookresearch/DPR}}for all the retrieval evaluations and MDR evaluation scripts \footnote{\url{https://github.com/facebookresearch/multihop_dense_retrieval}} for all the reader evaluations. 
\section{Computation}
Our \ourmodelshort has 182M paramteres. For \ourmodelshort pretraining, we use 32 V100-32GB GPUs, which takes about 3 days. For \ourmodelshort finetuning, we used 16 V100-32GB GPUs which takes about 2 days. Our reader model FiE has 330M parameters. We used 16 V100-32GB GPUs for training which takes about 1.5 days. For HotpotQA,
both the path reranker and the reader have 330M parameters. We used 16 V100-32GB GPUs for training, the path reranker takes about 12 hours and the reader takes about 4 hours to train. We train all of our models once due to the large computation cost.

\section{Licenses}
We list the License of the software and data used in this paper below:
\begin{itemize}
    \item DPR: CC-BY-NC 4.0 License
    \item MDR: CC-BY-NC 4.0 License
    \item Contriever: CC-BY-NC 4.0 License
    \item BLINK: MIT License
    \item NQ: CC-BY-SA 3.0 License
    \item HotpotQA: CC-BY-NC 4.0 License
    \item OTT-QA: MIT License 
    \item EntityQuestions: MIT License 
    \item SQuAD: CC-BY-SA 4.0 License
    \item WebQuestions: CC-BY 4.0 License 
\end{itemize}

\end{document}